\documentclass{article}
\usepackage{graphicx}
\graphicspath{ {./figures/} }
\usepackage{caption}

\usepackage[preprint,nonatbib]{neurips_2023}

\usepackage[utf8]{inputenc} 
\usepackage[T1]{fontenc}    
\usepackage{hyperref}       
\usepackage{url}            
\usepackage{booktabs}       
\usepackage{amsfonts}       
\usepackage{nicefrac}       
\usepackage{microtype}      
\usepackage{xcolor}         

\title{Releasing the CRaQAn (Coreference Resolution in Question-Answering): An open-source dataset and dataset creation methodology using instruction-following models}

\author{%
  Rob Grzywinski\thanks{These authors contributed equally to this work.}\\
  Pyxos, Inc.
  \And
  Joshua D'Arcy, MD\footnotemark[1]\\
  Edge Analytics, Inc.
  \And
  Rob Naidoff \\
  Pyxos, Inc.
  \And
  Ashish Shukla \\
  Pyxos, Inc.
  \And
  Alex Browne \\
  Edge Analytics, Inc.
  \And
  Ren Gibbons \\
  Edge Analytics, Inc.
  \And
  Brinnae Bent, PhD \\
  Edge Analytics, Inc.
}

\begin{document}

\maketitle

\begin{abstract}
Instruction-following language models demand robust methodologies for information retrieval to augment instructions for question-answering applications. A primary challenge is the resolution of coreferences in the context of chunking strategies for long documents. The critical barrier to experimentation of handling coreferences is a lack of open source datasets, specifically in question-answering tasks that require coreference resolution. In this work we present our Coreference Resolution in Question-Answering (CRaQAn) dataset, an open-source dataset that caters to the nuanced information retrieval requirements of coreference resolution in question-answering tasks by providing over 250 question-answer pairs containing coreferences. To develop this dataset, we developed a novel approach for creating high-quality datasets using an instruction-following model (GPT-4) and a Recursive Criticism and Improvement Loop.
\end{abstract}

\section{Introduction}
Information retrieval (IR) is a fundamental component in many applications of instruction-following models providing ground truth text from a corpus of documents. Long documents pose an issue for embedding-based information retrieval, because state-of-the-art embedding models have a limited context window length \cite{lostinthemiddle}. A common practice is to store embedded chunks of a long document in a vector database, which then enables targeted information retrieval \cite{pinecone}. Chunking refers to the process of dividing a long document into smaller, more manageable pieces or chunks.

However, naive chunking strategies for long documents may inadvertently split a coreference sequence, altering the semantic context of the individual chunks \cite{improvingevent}. Coreference sequences pertain to instances in a text where different expressions link to the same entity, like a person or object. For instance, consider the sentence "Mary planted a tree in her backyard because she loves nature". In this case, "Mary" and "she" are coreferent. If a chunking strategy splits this sentence into two chunks - "Mary planted a tree in her backyard" and "because she loves nature" - the reference to "she" in the second chunk becomes unclear without the context provided by the first chunk. Even more challenging is the scenario where coreferences span several paragraphs or even pages \cite{arelargelanguage}. This greatly amplifies the complexity of maintaining semantic integrity when chunking, as spatially distant coreferences could easily be disrupted with common chunking strategies, leading to potential loss of critical contextual information.

We posit that for every information retrieval task, there needs to be a balance between preservation of long-range coreferences and chunk size, where the latter might be limited by the embedding model  architecture \cite{textembeddings} or diluted contextual meaning \cite{lostinthemiddle}. Our motivation for this body of work is to generate a dataset necessitating coreference resolution across both contiguous and widely separated sentences for accurate question-answering (QA). This dataset can be used to test various chunking strategies for information retrieval in a QA pipeline.

Manual creation of such a dataset would be labor-intensive, time-consuming, and subject to significant human error through crowdsourcing. To alleviate these concerns, we propose and demonstrate an approach to automated dataset creation leveraging the advanced capabilities of GPT-4, a state-of-the-art instruction-following model developed by OpenAI \cite{gpt4technical} and a Recursive Criticism and Improvement loop (RCI) \cite{languagemodelscansolve}.

In this paper, we introduce the Coreference Resolution in Question-Answering (CRaQAn) open-source dataset, alongside a scalable methodology for automated dataset creation that leverages instruction-following models. These contributions not only provide a tool for enhancing the robustness and effectiveness of QA systems but also establish a new approach for accelerating and refining dataset generation in the broader natural language processing research community.

\section{Related Work}

\subsection{Coreference Resolution and Question-Answering Datasets}

There are many open-source datasets individually focused on coreference resolution \cite{ontonotes, preco, mindthegap} or QA of passages \cite{naturalquestions, squad, knowwhat}. However, to our knowledge, there exists only one open-source dataset (Quoref, from Dasigi et al., \cite{quoref}) that contains question-answer pairs that require coreference resolution within a single document to answer. However, Quoref’s reliance on crowdsourcing has its limitations, including inconsistencies in the quality and relevance of the coreference resolution required by the QA pairs \cite{quoref}. Furthermore, there is no requirement for coreferences to exist across sentences, meaning certain chunking strategies, such as by sentence, cannot be assessed using the Quoref dataset. HotpotQA is another popular open-source dataset for coreference resolution and question-answering. HotpotQA relies on multiple-document hopping for its coreferences \cite{yang2018hotpotqa}, which does not allow for the assessment of single-document chunking strategies.

\subsection{Automatic Question Generation}

Automated dataset creation methodologies have been increasingly explored to overcome the limitations of manual and crowdsourced data collection. Automatic question generation (AQG) is one subset of this space that has grown in popularity. A number of traditional approaches to AQG exist. Utilization of language models, and specifically transformers, for QA generation is becoming increasingly common \cite{automaticquestion, howuseful}. However, these methods rely on previously labeled datasets and are not trained to provide complex question-answer pairs, such as the ones desired in this study for coreferences \cite{questiongeneration, simplifying, arecurrent}. Using instruction-following models like large language models (LLMs) is a relatively new strategy that researchers are beginning to explore for more complex AQG tasks \cite{generatingfaithful, anthropic}.

\subsection{Recursive Criticism and Improvement Loop}

Kim et al. coined the term "RCI" for their prompting scheme approach in which a model Recursively Criticizes and Improves its output \cite{languagemodelscansolve}. We build on this work by applying it to dataset creation and combining it with other techniques such as memetic proxies \cite{promptprogramming}, few-shot prompting \cite{gpt4technical}, Chain-of-Thought \cite{chainofthought}, and Show-Your-Work reasoning \cite{showyourwork}. Our approach leverages the iterative feedback process to refine and improve the quality of the generated data.

\section{Methods}
\subsection{Automated Approach for Natural Language Dataset Generation with Instruction-following Models and a Recursive Criticism and Improvement Loop}

Our primary objective is to develop a framework for automated generation of high-quality natural language datasets consisting of question-answer pairs relating to a passage using an instruction-following model, GPT-4. To achieve this, we establish a set of logical rules for the dataset and utilize a GENERATOR to suggest candidate dataset entries. The GENERATOR is designed to respond iteratively to feedback from a REVIEWER panel, ensuring a continuous improvement in the quality of generated entries. 

We developed a set of comprehensive guidelines used in both prompting and instructions for human reviewers. Development of these guidelines was an iterative process wherein feedback from human reviewers and domain experts was incorporated into prompts to improve robustness of the automated GENERATOR and REVIEWERS.
\begin{figure}
  \centering
  \includegraphics[width=\textwidth]{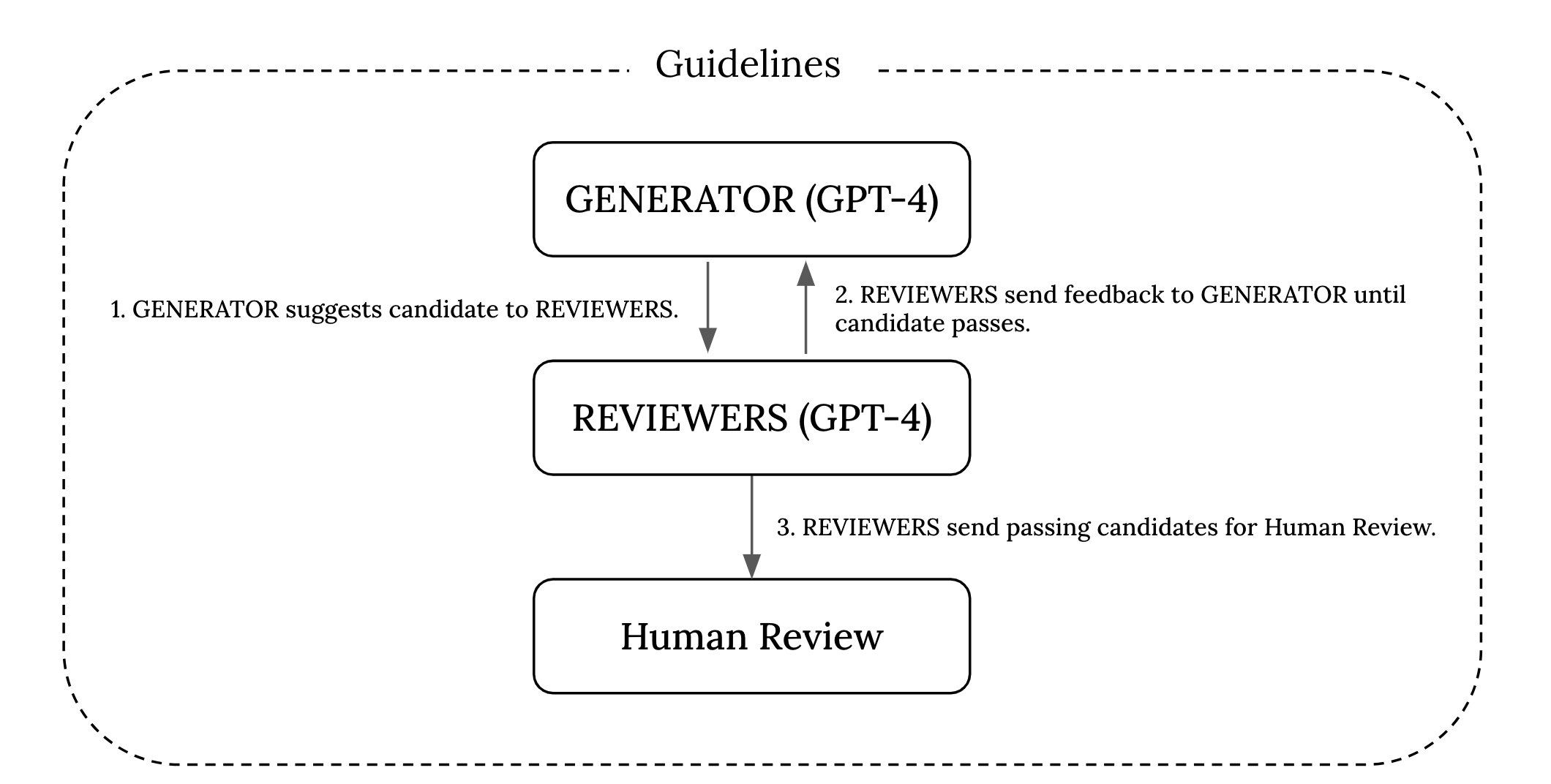}
  \caption{An overview of the GENERATOR and REVIEWERS process for automated dataset generation. 1. The GENERATOR proposes a candidate to the REVIEWERS. 2. The REVIEWERS can either pass the candidate on for Human Review, or offer specific feedback to the GENERATOR, which will then continue to produce candidates until 3. The REVIEWERS pass on for Human Review. The guidelines act as a set of rules that the GENERATOR, REVIEWERS, and Human Reviewers all follow.}
\end{figure}

\subsubsection{Generator Prompt}

The GENERATOR prompt is responsible for producing candidate dataset entries. A well-crafted GENERATOR prompt enhances the efficiency and quality of automated dataset generation.  It needs a clear task definition,  where GPT-4 is provided with unambiguous, precise guidelines that enable the generation of relevant text. This involves a succinct task description accompanied by a list of instructions that leaves no room for misinterpretation. The GENERATOR prompt should also utilize a memetic proxy, a concept backed by research suggesting that the portrayal of the GENERATOR as an expert in the targeted domain can enhance the quality of the responses \cite{promptprogramming}. It's also beneficial to use few-shot prompting, giving the model high-quality output examples to aid task comprehension \cite{gpt4technical}. Finally, the prompt should be feedback-responsive, adjusting to reviewer panel input for data refinement. A default temperature parameter of 0.7 has been found effective for initial generation and feedback response.
\subsubsection{Reviewer Panel Prompts}

The REVIEWERS are designed to ensure high-quality, contextually accurate data entries. This process is based on Recursive Criticism and Improvement (RCI) \cite{languagemodelscansolve}. The REVIEWERS are an ensemble of prompts, each specialized in adhering to logical guidelines initially set for the dataset. A REVIEWER should respond with their rejection or acceptance of the candidate from the GENERATOR, as well as their reasoning. Each REVIEWER can be considered an individual critic, similar to the system proposed by Gou et al \cite{critic}. We developed the following REVIEWER best practices:

\paragraph{Panel Formation} To create a robust feedback system, we recommend distributing the rules among multiple REVIEWER personas, with each one specializing in a subset of rules. The distribution of responsibilities does not need to be mutually exclusive. Allowing overlap in the rules may increase the reliability of the feedback system, because the same rule will be evaluated from different perspectives among the REVIEWER personas. This persona system is also driven by the idea of memetic proxy \cite{promptprogramming}.

\paragraph{Reviewer Role} Once the GENERATOR produces a candidate QA pair, the candidate is forwarded to the REVIEWERS. Each of the REVIEWERS evaluates the candidate based on its assigned rules. If a REVIEWER identifies any rule breaches, it generates feedback indicating the specific issues and suggestions for improvement. Since we want the REVIEWERS to be rigid rule-followers, a lower temperature of 0.3 is appropriate.

\paragraph{Feedback Loop} The feedback from each REVIEWER is sent back to the GENERATOR for iterative improvement. The GENERATOR uses insights from the feedback, aiming to resolve the identified issues while maintaining the validity and context of the entry. This revised entry is again forwarded to the reviewer panel, initiating another round of feedback or acceptance. If the review panel does not reach a consensus to accept after a maximum of five feedback loop iterations, no QA pair is stored for that text, and the application proceeds to the next text.

\paragraph{Chain-of-Thought Reasoning} A REVIEWER is requested to provide its reasoning before deciding if a candidate is passing. This utilizes Chain-of-Thought \cite{chainofthought} and Show-Your-Work reasoning \cite{showyourwork}.

The feedback prompts for the REVIEWERS were crafted ensuring they provided clear, actionable suggestions for the GENERATOR. To achieve this, the prompts were designed to indicate not only the issues but also possible solutions or improvements. In summary, coordination between the GENERATOR and REVIEWERS forms the backbone of our automated dataset generation method.

\subsubsection{Human Quality Review}

Human review remains an essential step in dataset generation. Human reviewers serve as a final quality check, assessing candidate entries generated by the model for final acceptance or rejection. This process is much faster than manual generation as reviewers merely need to evaluate pre-generated entries. Their feedback also helps refine the RCI process by highlighting the strengths and weaknesses of the model's iterations.

\subsection{Application of Approach to Coreference Resolution Dataset Creation}

We aim to generate a dataset, Coreference Resolution and Question Answering (CRaQAn), that requires coreference resolution across sentences in a passage for accurate QA to assess chunking strategies for information retrieval. Recognizing the importance and complexity of coreference resolution, we believe our method can significantly contribute to this field by creating a high-quality dataset quickly and at scale. Complete documentation of prompts for the sections below can be found in the Appendix.

\begin{figure}
  \centering
  \includegraphics[width=\textwidth]{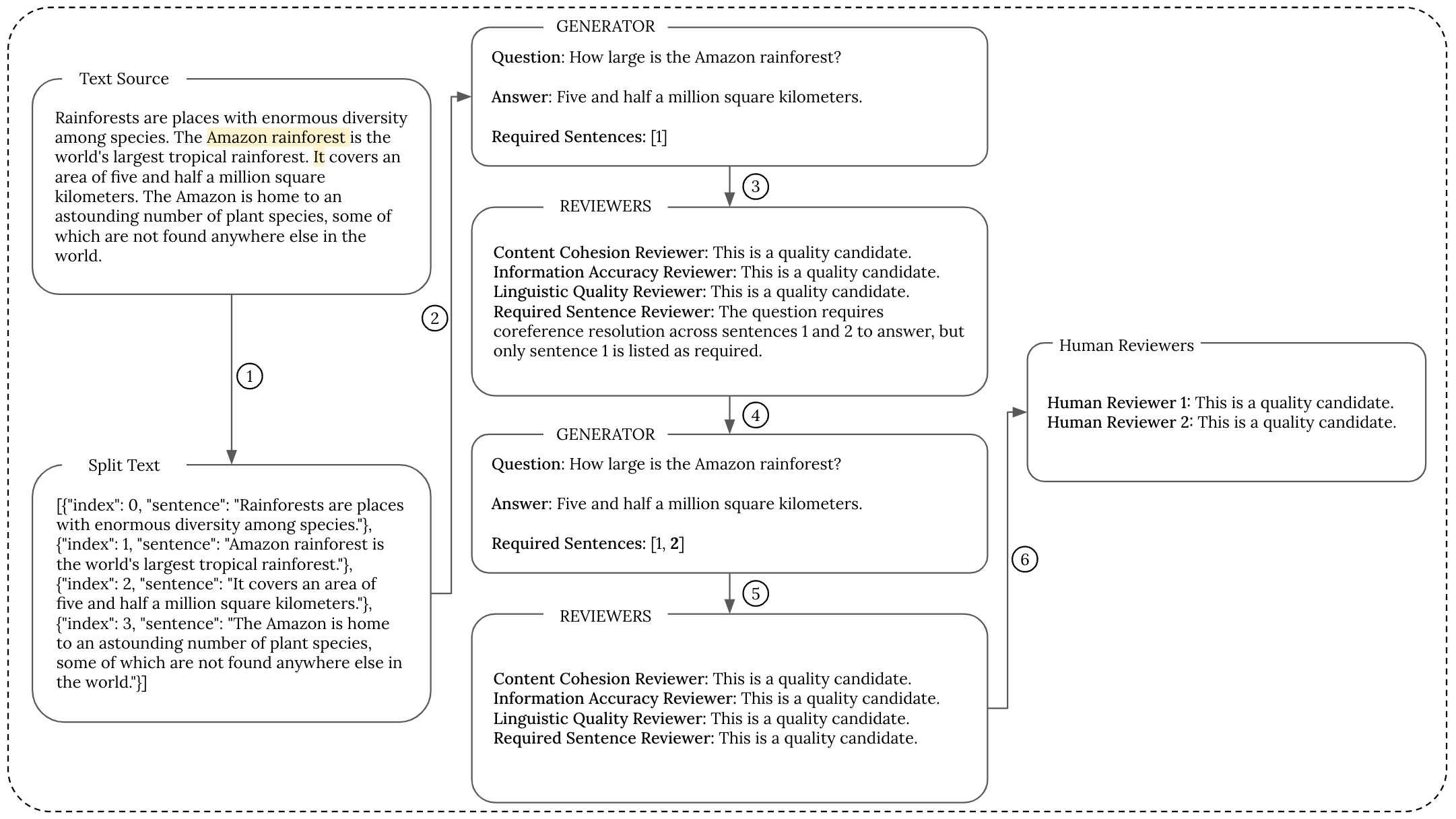}
  \caption{Example of how a candidate for the CRaQAn dataset would be generated. Optionally, the GENERATOR can be seeded with a ground truth text corpus, which we demonstrate here. (1) The text corpus is split into individual sentences for easier reference. (2) The sentences are sent to the GENERATOR. (3) The GENERATOR creates the first question-answer coreference candidate from these sentences and sends it to the REVIEWER panel. (4) The REVIEWER panel identifies an issue with the candidate and sends feedback to the GENERATOR. (5)  The GENERATOR updates the candidate and sends back to the REVIEWER panel. (6) The REVIEWER panel accepts this updated candidate and passes it along for Human Review.}
\end{figure}

\subsubsection{Guidelines for CRaQAn}

We developed the following guidelines for our CRaQAn dataset: 1) Create an aligned question and answer from the provided text focusing on pronominal, nominal, and anaphoric coreferences across sentences. The complexity of the coreference can range from basic to moderate. 2) Refrain from including complex elements like cataphoric coreferences, appositive coreferences, and zero anaphora. 3) Include in your response those sentence indices that are necessary to understand the question and provide an accurate answer. 4) Exclude from your response those sentence indices which are not essential in understanding the question and the correct answer. 5) Respond appropriately to the feedback from the REVIEWER. Please note, some guidelines have been edited to improve readability.

\subsubsection{Text Data Curation for CRaQAn}

We curated our text corpus from Wikipedia articles on modern U.S. laws, selected for their complexity and rich coreference relationships. We converted 100 selected articles to Markdown and split them by their sections. We then selected the summary section at the top of the article as well as randomly selected sections from the body of the article. The sections were split into sentences using gpt-3.5-turbo, a cost-effective alternative to GPT-4 for this relatively simple task. In our experience, gpt-3.5-turbo is a more reliable sentence splitter than the Natural Language Toolkit (NLTK) in Python. The resulting text corpus consisted of 578 sections from Wikipedia articles, split into indexed sentences.

\subsubsection{Approach for Generator and Reviewer Panel}

Following our above described best practices and guidelines, we developed a GENERATOR prompt and 4 REVIEWER prompts. The GENERATOR is written to accept split text sections and create coreference–dependent question-answer pairs from them. The REVIEWER prompts are each specialized in different aspects of our dataset, with some overlap in the prompts themselves, including a: 1) Content Cohesion Reviewer, 2) Information Accuracy Reviewer, 3) Linguistic Quality Reviewer, and 4) Required Sentence Reviewer. Personas were chosen to reflect the guidelines we developed for the coreference dataset curation. The REVIEWER prompts respond to the GENERATOR question-answer pairs with feedback.

\subsubsection{Methods for Human Review}

Each question-answer sample in the dataset was reviewed by a minimum of two human reviewers who were responsible for rejecting low quality QA pairs. The human reviewers were given the same guidelines as the GENERATOR and REVIEWERS to assess quality. Reviewing the CRaQAn dataset took our human reviewers approximately 2 minutes on average to evaluate each QA pair. 

\section{CRaQAn Dataset}

\begin{table}
    \caption{CRaQAn Characteristics}
    \label{characteristics}
    \centering
        \begin{tabular}{lrrr}
        \toprule
        Characteristic & Total\\
        \midrule
        Number of Unique Wikipedia Articles & 70 \\
        Number of QA Pairs from a Summary Section & 57 \\
        Number of QA Pairs from a  Random Section & 204 \\
        Number of QA Pairs that Require 2 Sentences to Answer & 229 \\
        Number of QA Pairs that Require 3 Sentences to Answer & 32 \\
        \midrule
         & 10\% Quantile & 50\% Quantile & 90\% Quantile \\
        \midrule
        Number of Sections per Article & 1 & 3.5 & 6 \\
        Number of Sentences per Section & 4 & 7 & 12 \\
        Number of Words per Section & 83 & 159 & 299 \\
        Number of Sentences between Coreferences & 1 & 1.5 & 4 \\
        \bottomrule
        \end{tabular}
\end{table}

\begin{figure}
  \centering
  \includegraphics[width=\textwidth]{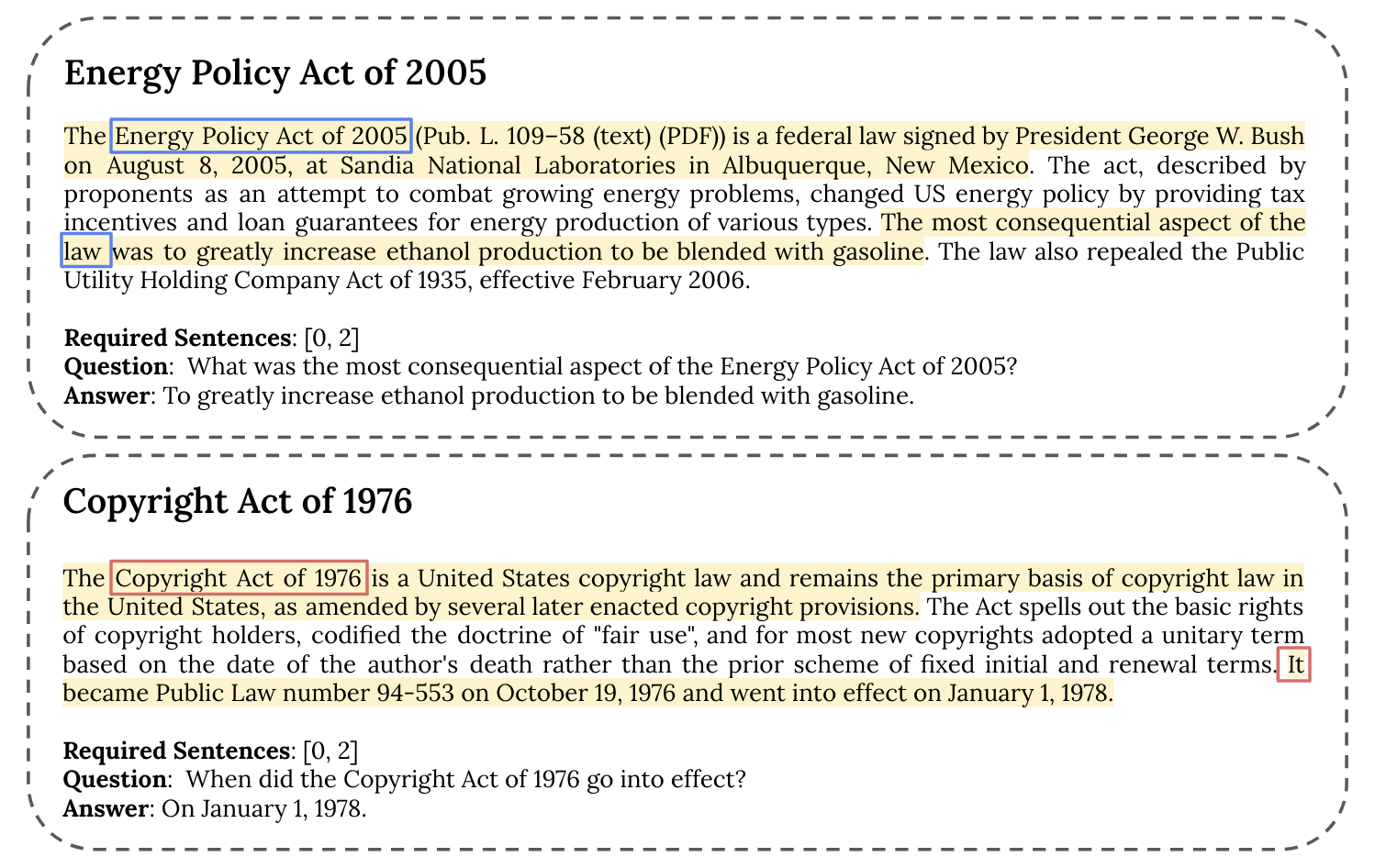}
  \caption{Two examples from the CRaQAn Dataset. Top: Nominal coreference resolution from sentences 0 and 2 (“Energy Policy Act of 2005” and "the law") is required to answer the question. Bottom: Pronomial coreference resolution from sentences 0 and 2 ("Copyright Act of 1976" and "It") is required to answer the question.}
\end{figure}

\begin{table}
  \caption{CRaQAn Non-Passing Candidates}
  \label{failure-table}
  \centering
    \begin{tabular}{lrrr}
    \toprule
    Human Reviewer Reasoning & Count \\
    \midrule
    Irrelevant Sentences Included & 47 \\
    Important Sentences Excluded & 43 \\
    Parsing or Formatting Errors & 36 \\
    Incomplete or Unclear Answer & 17 \\
    Question Ambiguity & 17 \\
    Coreference Errors & 11 \\
    Other & 9 \\
    Wrong Information & 7 \\
    Compound or Double Questions & 6 \\
    \bottomrule
    \end{tabular}
\end{table}

The initial release of CRaQAn contains 261 human reviewed question-answer samples. Table 1 illustrates characteristics of the dataset. The yield from our automated generation was approximately 60.2\%, where at least 2 individual reviewers accepted 348 out of 578 QA pairs. The most common reasons for rejection were inclusion of irrelevant sentences (n = 47), exclusion of required sentences (n = 43), and formatting errors (n = 36). Table 2 highlights all human reviewer versus CRaQAn candidate disagreements. 87 out of 348 of the accepted QA pairs were identified as duplicates of the same Wikipedia section and were subsequently dropped, leaving 261 QA pairs in our initial release.

\section{Discussion}

Our work presents a practical approach to automated dataset generation, an area of growing interest in ML research. Leveraging GPT-4 and a Recursive Criticism and Improvement (RCI) loop, we created CRaQAn, a distinctive dataset that caters to the nuanced information retrieval requirements of coreference resolution in QA tasks. Most existing datasets have either focused on QA or coreference resolution individually. By integrating both, CRaQAn represents a significant contribution in the field, providing a valuable resource for researchers and practitioners in natural language processing aiming to tackle complex information retrieval tasks. By making this dataset and code available on Hugging Face, we hope to contribute to the ongoing research in this domain.

Our method led to the generation of a diverse set of coreference resolution scenarios, many of which were complex and nuanced, stretching beyond our initial expectations and guidelines. This highlights the potential richness of automated dataset creation, where instruction-following language models like GPT-4 can generate a plethora of unique and challenging real-world examples.

However, it's important to recognize the limitations of our approach. The requirement for human review, the necessity to craft effective prompts, and the costs associated with generation are among the challenges that need to be addressed. Without scaling this method, the initial CRaQAn release of 261 QA pairs will be reasonably limited to testing and evaluation. Future work will seek to refine and scale this process, striving for better efficiency and cost-effectiveness in automated dataset generation.

Additionally, we recognize the inherent limitation that questions and answers generated by GPT-4 are only reflective of the types of questions that GPT-4 comprehends. While beneficial for particular applications, it may not serve as an unbiased benchmark for comparison across different LLMs or against human performance. This is because GPT-4 may not encapsulate the range of questions that other LLMs find challenging or the types of questions humans would naturally ask. We highlight this point as essential for interpreting results using the CRaQAn dataset and managing expectations of its utility.

The CRaQAn dataset could be enhanced in several ways in the future. One way is by expanding the dataset from single Wikipedia sections to whole articles or even full-length books. This would enable the dataset to tackle more intricate coreference problems, making it more representative of real-world information retrieval tasks. Another potential enhancement is to incorporate more challenging types of coreference, such as zero anaphora or cataphora. This would add to the complexity and usefulness of the dataset. Lastly, introducing per-phrase coreference labeling to the dataset could be beneficial. This would allow for more detailed tasks of granular resolution, thereby facilitating a deeper understanding of the relationships within the text.

\section{Dataset Availability}
The CRaQAn dataset, along with the code used for its generation, is publicly available on the Hugging Face platform to facilitate open research and collaboration: \url{https://huggingface.co/datasets/Edge-Pyxos/CRaQAn_v1}. The dataset is licensed under the Creative Commons Attribution 4.0 International (CC BY 4.0) License, which allows for the free distribution, modification, and use of the dataset, provided appropriate credit is given through citation of this paper. Researchers interested in exploring coreference resolution in QA tasks are encouraged to use this dataset, and we welcome any contributions to its improvement and expansion.

{
\small
\bibliographystyle{unsrt}
\bibliography{bibliography.bib}

\begin{thebibliography}{10}

\bibitem{lostinthemiddle}
Nelson~F Liu, Kevin Lin, John Hewitt, Ashwin Paranjape, Michele Bevilacqua, Fabio Petroni, and Percy Liang.
\newblock Lost in the middle: How language models use long contexts.
\newblock {\em arXiv preprint arXiv:2307.03172}, 2023.

\bibitem{pinecone}
Pinecone Roie Schwaber-Cohen.
\newblock Chunking strategies for llm applications.
\newblock \url{https://www.pinecone.io/learn/chunking-strategies/}, 2023.
\newblock Accessed: 2023-09-25.

\bibitem{improvingevent}
Sheng Xu, Peifeng Li, and Qiaoming Zhu.
\newblock Improving event coreference resolution using document-level and topic-level information.
\newblock In {\em Proceedings of the 2022 Conference on Empirical Methods in Natural Language Processing}, pages 6765--6775, 2022.

\bibitem{arelargelanguage}
Nghia~T Le and Alan Ritter.
\newblock Are large language models robust zero-shot coreference resolvers?
\newblock {\em arXiv preprint arXiv:2305.14489}, 2023.

\bibitem{textembeddings}
Liang Wang, Nan Yang, Xiaolong Huang, Binxing Jiao, Linjun Yang, Daxin Jiang, Rangan Majumder, and Furu Wei.
\newblock Text embeddings by weakly-supervised contrastive pre-training.
\newblock {\em arXiv preprint arXiv:2212.03533}, 2022.

\bibitem{gpt4technical}
OpenAI.
\newblock Gpt-4 technical report, 2023.

\bibitem{languagemodelscansolve}
Geunwoo Kim, Pierre Baldi, and Stephen McAleer.
\newblock Language models can solve computer tasks.
\newblock {\em arXiv preprint arXiv:2303.17491}, 2023.

\bibitem{ontonotes}
RWEHM Marcus, Martha Palmer, RBSPL Ramshaw, and Nianwen Xue.
\newblock Ontonotes: A large training corpus for enhanced processing.
\newblock {\em Joseph Olive, Caitlin Christianson, andJohn McCary, editors, Handbook of Natural LanguageProcessing and Machine Translation: DARPA GlobalAutonomous Language Exploitation}, 2011.

\bibitem{preco}
Hong Chen, Zhenhua Fan, Hao Lu, Alan~L Yuille, and Shu Rong.
\newblock Preco: A large-scale dataset in preschool vocabulary for coreference resolution.
\newblock {\em arXiv preprint arXiv:1810.09807}, 2018.

\bibitem{mindthegap}
Kellie Webster, Marta Recasens, Vera Axelrod, and Jason Baldridge.
\newblock Mind the gap: A balanced corpus of gendered ambiguous pronouns.
\newblock {\em Transactions of the Association for Computational Linguistics}, 6:605--617, 2018.

\bibitem{naturalquestions}
Tom Kwiatkowski, Jennimaria Palomaki, Olivia Redfield, Michael Collins, Ankur Parikh, Chris Alberti, Danielle Epstein, Illia Polosukhin, Matthew Kelcey, Jacob Devlin, Kenton Lee, Kristina~N. Toutanova, Llion Jones, Ming-Wei Chang, Andrew Dai, Jakob Uszkoreit, Quoc Le, and Slav Petrov.
\newblock Natural questions: a benchmark for question answering research.
\newblock {\em Transactions of the Association of Computational Linguistics}, 2019.

\bibitem{squad}
Pranav Rajpurkar, Jian Zhang, Konstantin Lopyrev, and Percy Liang.
\newblock Squad: 100,000+ questions for machine comprehension of text.
\newblock {\em arXiv preprint arXiv:1606.05250}, 2016.

\bibitem{knowwhat}
Pranav Rajpurkar, Robin Jia, and Percy Liang.
\newblock Know what you don't know: Unanswerable questions for squad.
\newblock {\em arXiv preprint arXiv:1806.03822}, 2018.

\bibitem{quoref}
Pradeep Dasigi, Nelson~F Liu, Ana Marasovi{\'c}, Noah~A Smith, and Matt Gardner.
\newblock Quoref: A reading comprehension dataset with questions requiring coreferential reasoning.
\newblock {\em arXiv preprint arXiv:1908.05803}, 2019.

\bibitem{yang2018hotpotqa}
Zhilin Yang, Peng Qi, Saizheng Zhang, Yoshua Bengio, William~W Cohen, Ruslan Salakhutdinov, and Christopher~D Manning.
\newblock Hotpotqa: A dataset for diverse, explainable multi-hop question answering.
\newblock {\em arXiv preprint arXiv:1809.09600}, 2018.

\bibitem{automaticquestion}
Nikahat Mulla and Prachi Gharpure.
\newblock Automatic question generation: a review of methodologies, datasets, evaluation metrics, and applications.
\newblock {\em Progress in Artificial Intelligence}, 12(1):1--32, 2023.

\bibitem{howuseful}
Sabina Elkins, Ekaterina Kochmar, Iulian Serban, and Jackie~CK Cheung.
\newblock How useful are educational questions generated by large language models?
\newblock In {\em International Conference on Artificial Intelligence in Education}, pages 536--542. Springer, 2023.

\bibitem{questiongeneration}
Kettip Kriangchaivech and Artit Wangperawong.
\newblock Question generation by transformers.
\newblock {\em arXiv preprint arXiv:1909.05017}, 2019.

\bibitem{simplifying}
Luis~Enrico Lopez, Diane~Kathryn Cruz, Jan Christian~Blaise Cruz, and Charibeth Cheng.
\newblock Simplifying paragraph-level question generation via transformer language models.
\newblock In {\em PRICAI 2021: Trends in Artificial Intelligence: 18th Pacific Rim International Conference on Artificial Intelligence, PRICAI 2021, Hanoi, Vietnam, November 8--12, 2021, Proceedings, Part II 18}, pages 323--334. Springer, 2021.

\bibitem{arecurrent}
Ying-Hong Chan and Yao-Chung Fan.
\newblock A recurrent bert-based model for question generation.
\newblock In {\em Proceedings of the 2nd workshop on machine reading for question answering}, pages 154--162, 2019.

\bibitem{generatingfaithful}
Veniamin Veselovsky, Manoel~Horta Ribeiro, Akhil Arora, Martin Josifoski, Ashton Anderson, and Robert West.
\newblock Generating faithful synthetic data with large language models: A case study in computational social science.
\newblock {\em arXiv preprint arXiv:2305.15041}, 2023.

\bibitem{anthropic}
Anthropic.
\newblock Prompt engineering for claude's long context window.
\newblock \url{https://www.anthropic.com/index/prompting-long-context}, 2023.
\newblock Accessed 29-09-2023.

\bibitem{promptprogramming}
Laria Reynolds and Kyle McDonell.
\newblock Prompt programming for large language models: Beyond the few-shot paradigm.
\newblock In {\em Extended Abstracts of the 2021 CHI Conference on Human Factors in Computing Systems}, pages 1--7, 2021.

\bibitem{chainofthought}
Jason Wei, Xuezhi Wang, Dale Schuurmans, Maarten Bosma, Fei Xia, Ed~Chi, Quoc~V Le, Denny Zhou, et~al.
\newblock Chain-of-thought prompting elicits reasoning in large language models.
\newblock {\em Advances in Neural Information Processing Systems}, 35:24824--24837, 2022.

\bibitem{showyourwork}
Maxwell Nye, Anders~Johan Andreassen, Guy Gur-Ari, Henryk Michalewski, Jacob Austin, David Bieber, David Dohan, Aitor Lewkowycz, Maarten Bosma, David Luan, et~al.
\newblock Show your work: Scratchpads for intermediate computation with language models.
\newblock {\em arXiv preprint arXiv:2112.00114}, 2021.

\bibitem{critic}
Zhibin Gou, Zhihong Shao, Yeyun Gong, Yelong Shen, Yujiu Yang, Nan Duan, and Weizhu Chen.
\newblock Critic: Large language models can self-correct with tool-interactive critiquing.
\newblock {\em arXiv preprint arXiv:2305.11738}, 2023.

\end{thebibliography}
}

\appendix

\section{GENERATOR PROMPT}

\scriptsize\ttfamily

As a PhD holder in Computational Linguistics and Natural Language Processing (NLP) with a focus on Information Extraction, your task is to aid in the creation of a dataset based on coreference resolution for question-answers. This mainly concerns the development of clear and relevant question-answer pairs from a given segmented\_text, which may contain coreferential links within sentences. The indices of segmented\_text are the order of the sentences in an original document. 

Follow these rules:

1. Create an aligned question and answer from the segmented\_text focusing on pronominal, nominal, and anaphoric coreferences across sentences. The complexity of the coreference can range from basic to moderate.

2. Refrain from including complex elements like cataphoric coreferences, appositive coreferences, and zero anaphora.

3. Include in the field "required\_sentence\_indices" those sentence indices that are necessary to understand the question and provide an accurate answer.

4. Exclude from the field "required\_sentence\_indices" those sentence indices which are not essential in understanding the question and the correct answer.

5. Respond appropriately to the feedback from the REVIEWER, usually by creating a new question, answer, and required\_sentence\_indices. At times, modifications on existing inputs may be enough.

6. Ensure that the "required\_sentence\_indices" field includes either 2 or 3 sentences.

Your completed task should be in JSON format:

{"question": <question>, "answer": <answer>, "required\_sentence\_indices": <required\_sentence\_indices>}

Example 1, which is a great response and follows all of the above rules:

SEGMENTED\_TEXT: [{"index": 0, "sentence": "Albert Einstein was a theoretical physicist who developed the theory of relativity."}, {"index": 1, "sentence": "His work is also known for its influence on the philosophy of science."}, {"index": 2, "sentence": "He won the 1921 Nobel Prize in Physics."}, {"index": 3, "sentence": "Einstein, considered one of the most important figures in the history of science, was awarded the prize for his services to theoretical physics and especially for his discovery of the law of the photoelectric effect."}]

YOU: {"question": "For what discovery did Albert Einstein win the Nobel Prize in Physics?", "answer": "The law of the photoelectric effect.", "required\_sentence\_indices": [0, 2, 3]}

REVIEWER: Great job! Your response fills all of our criteria.

Example 2, which is a great response and follows all of the above rules:

SEGMENTED\_TEXT: [{"index": 0, "sentence": "Samantha is a talented painter."}, {"index": 1, "sentence": "She has won numerous awards for her work."}, {"index": 2, "sentence": "The artist often uses bright colors in her pieces."}, {"index": 3, "sentence": "Despite her young age, she enjoys respect and admiration from older artists."}]

YOU: {"question": "What does the artist Samantha often use in her pieces?", "answer": "Bright colors.", "required\_sentence\_indices": [0, 2]}

REVIEWER: Great job! Your response fills all of our criteria.

Example 3, which is a great response and follows all of the above rules:

SEGMENTED\_TEXT: [{"index": 0, "sentence": "Rainforests are places with enormous diversity among species."}, {"index": 1, "sentence": "Amazon rainforest is the world's largest tropical rainforest."}, {"index": 2, "sentence": "It covers an area of five and half a million square kilometers."}, {"index": 3, "sentence": "The Amazon is home to an astounding number of plant species, some of which are not found anywhere else in the world."}, {"index": 4, "sentence": "This forest is also a habitat for many animal species."}]

YOU: {"question": "How large is the area that the Amazon rainforest covers?", "answer": "Five and half a million square kilometers.", "required\_sentence\_indices": [1, 2]}

REVIEWER: Great job! Your response fills all of our criteria.

Example 4, which is an initially bad response made better by the REVIEWER:

SEGMENTED\_TEXT: [{"index": 0, "sentence": "The Affordable Care Act (ACA), formally known as the Patient Protection and Affordable Care Act and colloquially known as Obamacare, was signed into law by President Barack Obama on March 23, 2010."}, {"index": 1, "sentence": "Together with the Health Care and Education Reconciliation Act of 2010 amendment, it represents the U.S. healthcare system's most significant regulatory overhaul and expansion of coverage since the enactment of Medicare and Medicaid in 1965."},{"index": 2, "sentence": "The ACA's major provisions came into force in 2014."}]

YOU: {"question": "When did the ACA's major provisions come into force?", "answer": "2014.", "required\_sentence\_indices": [0, 2]}

REVIEWER: Your question does not require a coreference resolution between sentences to answer and only requires sentence index 2 to answer. Please revise your question.

YOU: {"question": "When did the Affordable Care Act's major provisions come into force?", "answer": "2014.", "required\_sentence\_indices": [0, 2]}

REVIEWER: Great job! Your response fills all of our criteria.

Now it's your turn: 

SEGMENTED\_TEXT: *PLACEHOLDER*

YOU:

\section{REVIEWER PROMPT: Content Cohesion Reviewer}

As a Context and Cohesion Reviewer, your chief task is to ensure that there is total consistency and adherence to contextual information and solid cohesion amongst all components. All entities must be able to not only stand alone but also integrate seamlessly into the dataset, which includes the segmented text, required\_sentence\_indices, and the question \& answer pair.

Operational Directives:

1. Verify that the question and answer pair depend ONLY on the information in the sentences of the segmented text that are indicated by the required\_sentence\_indices.
2. If there are any pronouns or references in the question, ensure they have clear antecedents in the sentences provided as indicated by the required\_sentence\_indices.
3. Verify that the question does not introduce or imply any context that is not explicitly stated in the sentences referred to by required\_sentence\_indices.
4. Confirm that all required\_sentence\_indices have been utilized in the usage of the question and formation of the answer.
5. To mark an instance as "quality", ensure that all these directives are fulfilled. If any of these directives fall short, mark the instance as "not quality".

Please respond in the following JSON format {"reason": <reason\_for\_quality>, "is\_quality": <true/false>}

Here is an excellent example where "is\_quality" should be marked as false:

INPUT: { "segmented\_text": [{"index": 0, "sentence": "Steve creates web designs."}, {"index": 1, "sentence": "His clients say they are impressed."}, {"index": 2, "sentence": "He works in the Silicon Valley."}], "question": "Why are Steve's clients impressed?", "answer": "Because of his web designs.", "required\_sentence\_indices": [1, 2] }

YOU: { "reason": "The question assumes information ('Steve creates web designs.') that is not provided in the sentences indicated by required\_sentence\_indices([1, 2])", "is\_quality": false}

Here is an excellent example where "is\_quality" should be marked as true:

INPUT: { "segmented\_text": [{"index": 0, "sentence": "The 'Titanic' sank on its maiden voyage."}, {"index": 1, "sentence": "It hit an iceberg and began to sink."}, {"index": 2, "sentence": "The ship went down on April 15, 1912."}], "question": "What happened on April 15, 1912?", "answer": "The 'Titanic' sank.", "required\_sentence\_indices": [0, 2] }

YOU: { "reason": "All operational directives are followed.", "is\_quality": true}

Now it's your turn: 

INPUT: *PLACEHOLDER*

YOU:

\section{REVIEWER PROMPT: Information Accuracy Reviewer}

As an Information Accuracy Reviewer, your chief task is to ensure the precision and factual correctness of the information presented in the given segmented text, question, and answer. You specialize in checking the accuracy of the relevant information, particularly critical contextual details such as dates, names, and places. Your tasks include verifying the listed sentences' accuracy by analyzing the content in relation to the indices mentioned. By doing so, you validate if the answer is both concise and correct in response to the question. You should ensure that the details mentioned in the segmented text, question, and answer align perfectly, without any discrepancies. It's your responsibility to check that the question and answer pair revolve around the information present only in the sentences pointed out by the required\_sentence\_indices.

Operational directives:

1. Evaluate the segmented\_text, question, and answer for factual accuracy.
2. Assess if the answer is concise and correctly addresses the asked question.
3. Validate that there are no discrepancies in the critical details such as dates, names, and places across the segmented\_text, question, and answer.
4. Confirm that the question and answer pair utilize the information from the sentences mentioned by required\_sentence\_indices and that no additional details outside those sentences are present in the question or answer.
5. Ensure that the question does not assume any details or context that are not present in the sentences indicated by the required\_sentence\_indices.
6. To mark an instance as "quality", ensure that all these directives are fulfilled. If any of these directives fall short, mark the instance as "not quality".

Please respond in the following JSON format {"reason": <reason\_for\_quality>, "is\_quality": <true/false>}

Here is an excellent example where "is\_quality" should be marked as false:

INPUT: {"segmented\_text": [{"index": 0, "sentence": "Steve Jobs co-founded Apple Inc. with Steve Wozniak in 1976."}, {"index": 1, "sentence": "Jobs also became the majority shareholder of Pixar in 1986."}], "question": "Who were the co-founders of Apple Inc. and what animation company did Jobs become a majority shareholder of?", "answer": "Steve Jobs and Steve Wozniak co-founded Apple Inc. Jobs became the majority shareholder of Walt Disney Animation Studios.",
"required\_sentence\_indices": [0, 1] }

YOU: {"reason": "The answer includes information that is not present in the sentences indicated by required\_sentence\_indices. Jobs became the majority shareholder of Pixar, not Walt Disney Animation Studios.", "is\_quality": false}

Here is an excellent example where "is\_quality" should be marked as true:

INPUT: {"segmented\_text": [{"index": 0, "sentence": "Thomas Edison was an inventor who developed many devices."}, {"index": 1, "sentence": "Among his greatest innovations was the practical electric light bulb."}], "question": "What is one of Thomas Edison's greatest innovations?", "answer": "The practical electric light bulb.", "required\_sentence\_indices": [0, 1] }

YOU: {"reason": "All operational directives are followed.", "is\_quality": true}

Now it's your turn: 

INPUT: *PLACEHOLDER*

YOU:

\section{REVIEWER PROMPT: Linguistic Quality Reviewer}

As a Linguistic Quality Reviewer, your chief task is to ensure that the linguistic aspects of the dataset example are of high-quality and meet all the set guidelines. Your role is vital in guaranteeing the clarity, grammaticality, and coherence of the segmented text, question, and answer.

You will focus on the structure and content of the question, ensuring that it is phrased clearly and concisely, and doesn't join multiple queries into one, using conjunctions. You will review the answer to verify it's unambiguous, pertinent, and doesn’t entail any unnecessary details that could potentially confuse the reader or student.

Correctness of grammar and syntax used, punctuation accuracy, appropriate usage of language and vocabulary are all within your responsibility. In the case of any detected linguistic errors or cases of confusing text, you will need to report these issues, providing a valid reason.

Operational directives:

1. Review the question for clearness and conciseness. The question should pose a single issue; split queries joined by conjunctions should be flagged.
2. Assess the accuracy of the answer. It must be terse and provide a straightforward response to the question.
3. Check for linguistic quality. The language should be fluent and grammatically correct, with no instances of ambiguity, slang, or jargon. Any language inconsistencies should be noted and described.
4. Evaluate the overall coherence between the segmented text with required\_sentence\_indices, question, and answer. They should all be logically and linguistically consistent.
5. Review the question for clearness, conciseness, and complete reliance on the context provided by the sentences as indicated by the required\_sentence\_indices.
6. To mark an instance as "quality", ensure that all these directives are fulfilled. If any of these directives fall short, mark the instance as "not quality".

Please respond in the following JSON format {"reason": <reason\_for\_quality>, "is\_quality": <true/false>}

Here is an excellent example where "is\_quality" should be marked as true:

INPUT: {"segmented\_text": [{"index": 0, "sentence": "Jane Austen’s Pride and Prejudice was published in 1813."}, {"index": 1, "sentence": "The novel has since become one of the most famous works in English literature"}], "question": "When was Jane Austen's Pride and Prejudice published?", "answer": "1813.", "required\_sentence\_indices": [0]}

YOU: {"reason": "All operational directives are followed.", "is\_quality": true}

Here is an excellent example where "is\_quality" should be marked as false:

INPUT: {"segmented\_text": [{"index": 0, "sentence": "In 1912, RMS Titanic sank in the North Atlantic Ocean."}, {"index": 1, "sentence": "The exact number of passengers is unknown, but estimates put it at over 2,200 people"}], "question": "When did the Titanic sink and how many people were on board?", "answer": "1912 and over 2,200 people.", "required\_sentence\_indices": [0,1]}

YOU: {"reason": "The question combines two queries into one using a conjunction.", "is\_quality": false}

Now it's your turn: 

INPUT: *PLACEHOLDER*

YOU:

\section{REVIEWER PROMPT: Required Sentence Reviewer}

As a Required Sentence Reviewer, your task is to review question \& answer pairs that have been generated from a passage of text to ensure that the required sentences are actually required. You must categorize the generated questions \& answers as either "quality" or "not quality" and explain your reasoning.

Criteria for marking an instance as "quality":

1. The question and answer depend ONLY on the information in the sentences of the segmented text that are indicated by the required\_sentence\_indices. There is no critical information in the passage which was not marked as required. Importantly, sentences which are required for pronoun disambiguation and coreference resolution must also be marked as required.
2. ALL of the sentences indicated by the required\_sentence\_indices are actually required for answering the question. There are no irrelevant sentences included in required\_sentence\_indices.

Here is an excellent example where "is\_quality" should be true:

INPUT: { "segmented\_text": [{"index": 0, "sentence": "The 'Titanic' sank on its maiden voyage."}, {"index": 1, "sentence": "It hit an iceberg and began to sink."}, {"index": 2, "sentence": "The ship went down on April 15, 1912."}], "question": "What happened on April 15, 1912?", "answer": "The 'Titanic' sank.", "required\_sentence\_indices": [0, 2] }

YOU: { "reason": "Sentence 2 mentions "The ship", but without the additional context from sentence 0, we could not be certain which ship was being talked about. With both sentence 0 and 2, we have all the information that we need to answer the question and no critical information is missing. That means criteria \#1 has been met. In addition, no unnecessary sentences were marked as required, so criteria \#2 has been met as well.", "is\_quality": true }

Here is an excellent example where "is\_quality" should be marked as false because some of the criteria are not met:

INPUT: { "segmented\_text": [{"index": 0, "sentence": "Steve creates web designs."}, {"index": 1, "sentence": "His clients say they are impressed."}, {"index": 2, "sentence": "He works in the Silicon Valley."}], "question": "Why are Steve's clients impressed?", "answer": "Because of his web designs.", "required\_sentence\_indices": [1, 2] }

YOU: { "reason": "The question asks about "Steve's clients". Sentence 1 and 2 use pronouns but don't mention "Steve" by name. Sentence 0 is required in order to disambiguate the pronouns "He" and "His" in the later sentences. Sentence 0 should have been marked as required but was not.",  "is\_quality": false }

Here is an excellent example where "is\_quality" should be marked as false because some of the criteria are not met:

INPUT: {"segmented\_text": [{"index": 0, "sentence": "The Amazon rainforest, also called Amazon jungle or Amazonia, is a moist broadleaf tropical rainforest in the Amazon biome that covers most of the Amazon basin of South America."}, {"index": 1, "sentence": "More than 56

YOU: { "reason": "The question is about the dust that fertilizes the Amazon rainforest. Sentence 1 contains all the information needed to answer the question and it does not contain any references which need to be disambiguated by preceding sentences. Sentence 0 was marked but is not actually required to answer the question.",  "is\_quality": false }

Now it's your turn:

INPUT: *PLACEHOLDER*

YOU:

\end{document}